%% file: main.tex
\documentclass[letterpaper, 10 pt, conference]{ieeeconf}  

\IEEEoverridecommandlockouts                              

\usepackage{graphics} 
\usepackage{graphicx}
\usepackage{amsmath} 
\usepackage{amssymb}  
\usepackage{booktabs}
\usepackage{xcolor, colortbl}
\usepackage{caption}
\usepackage{subcaption}
\usepackage{lipsum}
\usepackage{cuted}
\usepackage[font=small,labelfont=bf,tableposition=top]{caption}
\usepackage{cite}
\usepackage{microtype}

\title{\LARGE \bf
Density-aware NeRF Ensembles:\\Quantifying Predictive Uncertainty in Neural Radiance Fields%
}

\author{Niko S\"underhauf, Jad Abou-Chakra, Dimity Miller
\thanks{The authors are with Queensland University of Technology (QUT) in Brisbane, Australia, and  acknowledge the ongoing support of the QUT Centre for Robotics. Contact: {\tt\small niko.suenderhauf@qut.edu.au}}%
}
\input{mathstuff.tex}

\begin{document}

\maketitle

\thispagestyle{empty}

\pagestyle{empty}

\begin{abstract}
We show that \emph{ensembling} effectively quantifies model uncertainty in Neural Radiance Fields (NeRFs) if a density-aware epistemic uncertainty term is considered. 
The naive ensembles investigated in prior work simply average rendered RGB images to quantify the model uncertainty caused by conflicting explanations of the observed scene. In contrast, we additionally consider the termination probabilities along individual rays to identify epistemic model uncertainty due to a lack of knowledge about the parts of a scene \emph{unobserved} during training.
We achieve new state-of-the-art performance across established uncertainty quantification benchmarks for NeRFs, outperforming methods that require complex changes to the NeRF architecture and training regime.  
We furthermore demonstrate that NeRF uncertainty can be utilised for next-best view selection and model refinement. 
\end{abstract}

\section{Introduction}

Neural Radiance Fields~\cite{mildenhall2021nerf} (NeRFs) implicitly represent the geometry and appearance of complex 3D scenes as a continuous function that is implemented as a relatively simple deep neural network. NeRFs and other implicit representations were met with an immense interest in the past two years, leading to an often-quoted ``explosion'' of work in this area~\cite{dellaert2020neural}. While most of this work has been conducted by the computer graphics and vision communities, researchers in robotics have quickly started to explore possible use cases of NeRFs for important robotics tasks such as navigation~\cite{adamkiewicz2022navigation}, SLAM~\cite{sucar2021imap,zhu2022nice}, or manipulation~\cite{li2022visumotor, yen2022nerf, ichnowski2021dex}. Since the appearance of highly efficient NeRFs, such as Instant-NGP~\cite{mueller2022instant}, that can be trained in seconds rather than many hours, the adoption of NeRFs for robotics has become palpable.

NeRFs provide an interesting new take on the long-standing problem of how to represent the 3D world for robotics, with all its geometric and semantic complexity~\cite{cadena2016past,rosen2021advances,garg2020semantics}. However, NeRFs face the same challenges as other deep learning approaches in robotics~\cite{sunderhauf2018limits} -- they lack the ability to express uncertainty in their predictions. 

The incorporation of uncertainty and the generally probabilistic nature of data, estimations, and predictions is well-established in large parts of the robotics literature~\cite{thrun2002probabilistic}, yet it is still a highly active area of research in deep learning~\cite{abdar2021review}. Approaches range from  principled Bayesian Deep Learning~\cite{mackay1992practical, neal2012bayesian} to simple but surprisingly effective approximate methods such as MC Dropout~\cite{gal2016dropout} or Deep Ensembles~\cite{lakshminarayanan2017simple}. 
Despite the large body of work in NeRFs, little research has investigated adopting the above methods to quantify the predictive uncertainty of NeRFs. 

\begin{figure}[t]
    \centering
    \includegraphics[width=0.49\linewidth]{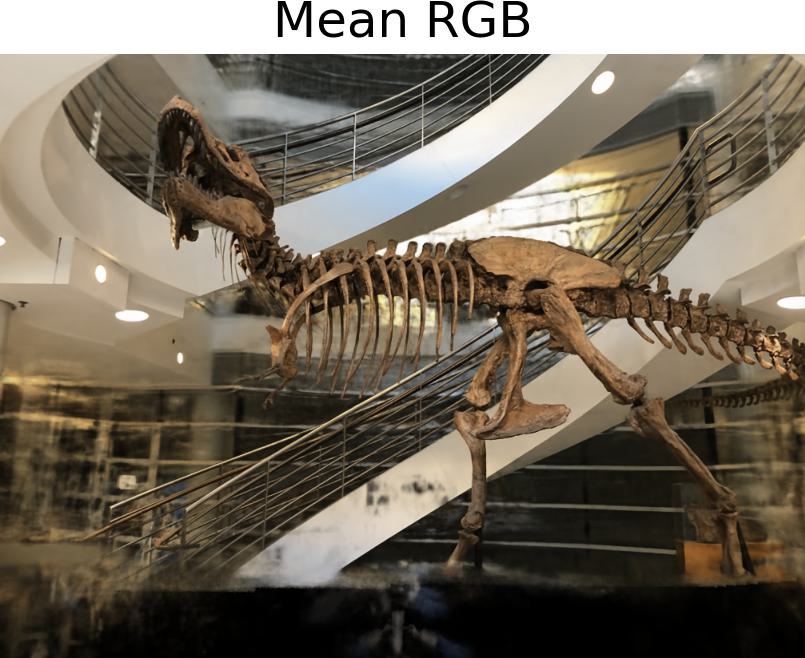}       
    \includegraphics[width=0.49\linewidth]{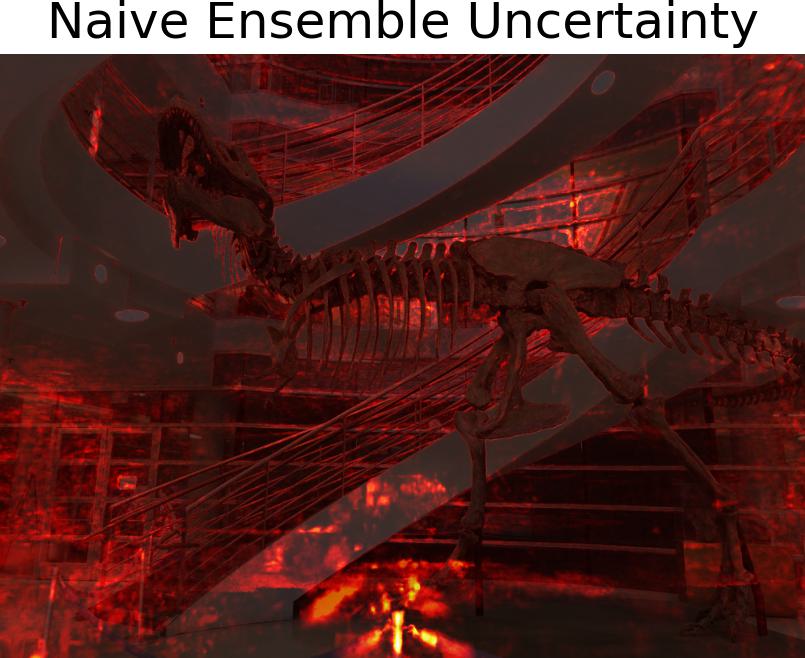}       
    \includegraphics[width=0.49\linewidth]{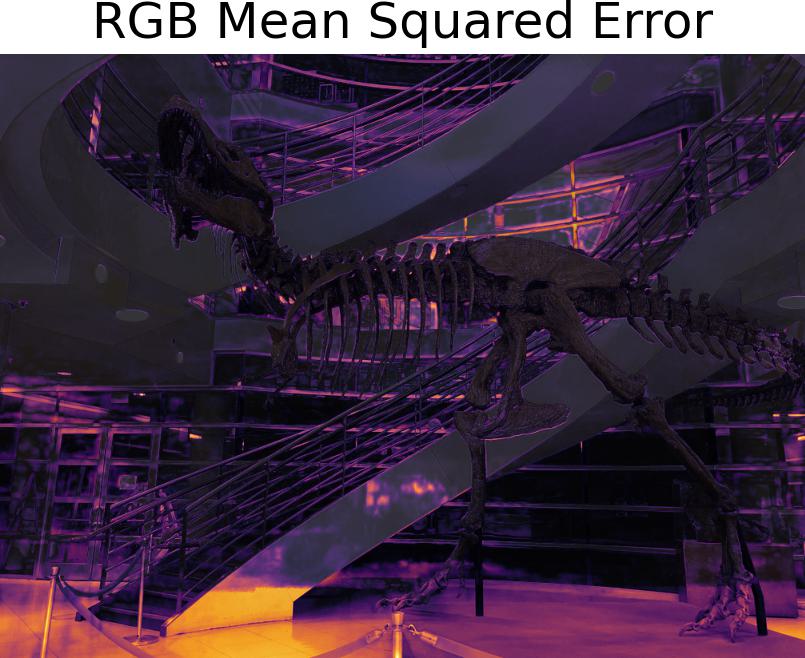}       
    \includegraphics[width=0.49\linewidth]{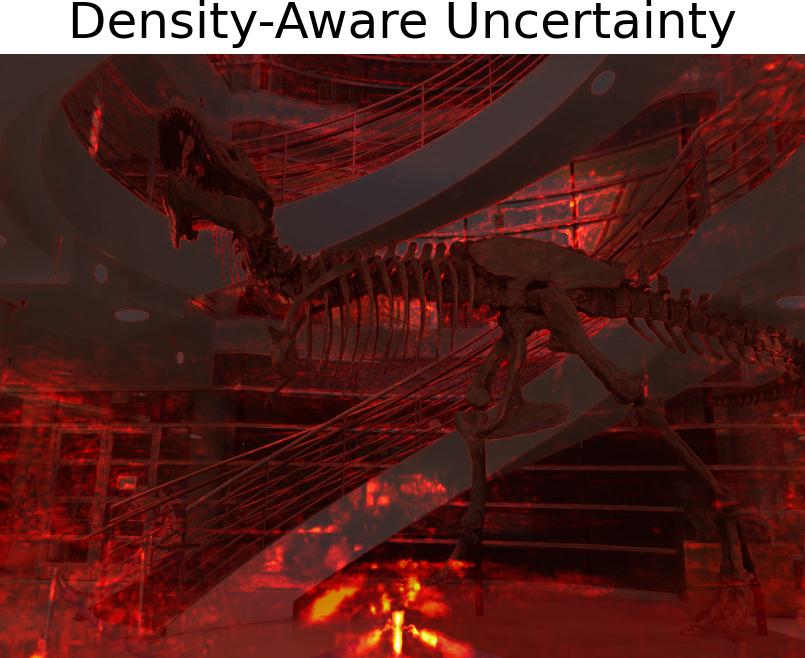}       
    \caption{An ensemble of Neural Radiance Fields can render a mean RGB image and use the colour variance in pixel space to quantify its predictive uncertainty (top).    
    However, this naive approach to ensembling often does not capture the epistemic uncertainty in parts of a scene that were \emph{unobserved} during training. In our example, the ensemble agrees to render the unobserved bottom portion of the images in black, resulting in negligible uncertainty, despite the high error (bottom left) when compared to the ground truth. We show that an epistemic uncertainty term that captures the termination probabilities along each ray must be considered in addition to the RGB variance to make ensembling an effective approach to quantifying uncertainty in Neural Radiance Fields (bottom right). }
    \label{fig:hero}
\end{figure}

Our paper addresses this gap. Specifically, we show that \emph{ensembling}~\cite{lakshminarayanan2017simple} is an effective approach to quantify uncertainty in Neural Radiance Fields and that prior work has dismissed the ensembles approach prematurely~\cite{shen2021stochastic, shen2022conditional}. Our key insight is that instead of only averaging and calculating variance in the RGB space, a NeRF ensemble should consider an additional epistemic uncertainty~\cite{kendall2017uncertainties} term that depends on the densities and termination probabilities along individual rays. We show that such a \emph{density-aware} ensemble of Instant-NGP NeRFs~\cite{mueller2022instant} achieves new state-of-the-art performance in terms of uncertainty quality, and can furthermore effectively select the next-best view when iteratively building a training dataset and refining an implicit object model.

\section{Related Work}

\subsection{Neural Radiance Fields}
Neural Radiance Fields~\cite{mildenhall2021nerf} learn a radiance and density field that -- in combination with a volumetric rendering process -- can explain a set of posed training images and realistically render novel views. Many variations of the original NeRF have emerged~\cite{xie2022neural}, including follow-up works~\cite{tancik2020fourfeat, Chen2022ECCV, yu_and_fridovichkeil2021plenoxels, barron2021mipnerf, mueller2022instant} that highlight the impact of the input encoding on training and inference speed. Instant-NGP~\cite{mueller2022instant} is one of the fastest and most optimised formulations, using a parametric encoding alongside a smaller MLP. It trains in seconds and renders at real-time rates, making it a great candidate for adoption in robotics, and thus is the backbone chosen for our work. NeRFs provide an exciting new approach to scene representation in robotics, and a variety of use cases are currently explored. For example, NiceSLAM~\cite{zhu2022nice} and iMap~\cite{sucar2021imap} use a NeRF to represent a map within a SLAM system, while~\cite{22-driess-NeRF-RL-preprint, li2022visumotor} use it as a decoder within an autoencoder framework to learn an alternative representation for planning and reinforcement learning, and~\cite{yen2022nerf, ichnowski2021dex} use NeRFs for data augmentation.


\subsection{Uncertainty Quantification in Deep Learning}
Uncertainty Quantification in Deep Learning is a rapidly growing field and we refer the reader to~\cite{abdar2021review} for a recent review. Kendall et al.~\cite{kendall2017uncertainties} identified two relevant types of uncertainty for computer vision -- aleatoric uncertainty and epistemic uncertainty. Aleatoric uncertainty refers to uncertainty in the input data, introduced by noise or random processes~\cite{kendall2017uncertainties}. In contrast, epistemic uncertainty refers to uncertainty in the model's learnt parameters, representing the model's lack of knowledge due to a finite training dataset \cite{kendall2017uncertainties}. Deep Ensembles is a popular approach for uncertainty quantification, producing predictive uncertainty that captures both aleatoric and epistemic uncertainty~\cite{lakshminarayanan2017simple}. 



\subsection{Uncertainty Quantification for NeRFs}

Stochastic NeRF (S-NeRF)~\cite{shen2021stochastic} is one of the few papers investigating uncertainty quantification for NeRFs. It reformulates the NeRF optimisation as a Bayesian estimation problem, and applies a Variational Inference approach to effectively estimate the posterior distribution over the parameters of all possible radiance fields given the observed training data. Mean and variance for each rendered pixel can be calculated by sampling from the approximated posterior distribution over all radiance fields. The variance in pixel space is then used as the uncertainty measure for a particular pixel. Conditional-Flow NeRF (CF-NeRF)~\cite{shen2022conditional} by the same authors builds on this work and relaxes some of S-NeRF's constraints on the involved distributions, especially the independence assumption between radiance and density. 

Both methods~\cite{shen2021stochastic,shen2022conditional} involve a complex reformulation of the NeRF architecture, rendering process, and its training regime, limiting the applicability of the proposed approach to other NeRF implementations such as Instant NGP~\cite{mueller2022instant}. In contrast, the ensembles approach we propose here is extremely simple to implement as it does not require any changes in the underlying NeRF architecture. While the research field around implicit representations continues to evolve rapidly and better or faster NeRF formulations are regularly proposed, we see the ability of easy adaptation as a main advantage of our method.


\input{uncertainty.tex}

\input{exploration.tex}

\section{Discussion, Conclusions, and Future Work}
We have shown that the key to making ensembling an effective approach for uncertainty quantification in NeRFs is to combine the simple RGB variance with an additional epistemic uncertainty term that is informed by the predicted densities along each individual ray. 

A major advantage of our method is the simplicity of adopting ensembling strategies to new emerging variants of NeRFs. This allows us to leverage progress regarding training time, required training views, representational power or rendering speed, while maintaining the ability to readily quantify predictive uncertainty in emerging NeRFs in the future. NeRF ensembling approaches are already computationally feasible after the appearance of hash-encoding NeRFs~\cite{mueller2022instant} that can train in mere seconds (with the option of training all ensemble members in parallel) and render in real time. 

Our qualitative results (Fig.~\ref{fig:rays}) suggest that there is no strict distinction between aleatoric and epistemic~\cite{kendall2017uncertainties} uncertainty in our uncertainty measure. Being able to separately quantifying both is worthwhile for future work, since in some applications (e.g. next-best view selection) epistemic uncertainty is more informative than aleatoric uncertainty.  

Another interesting direction for future work is to use our density-aware ensemble to guide exploration and navigation of a mobile robot through an unknown scene, while mapping it with the NeRF. While we have shown the uncertainty to be effective in selecting the next-best view from a set of candidate views, we are interested in extracting the gradient of the uncertainty measure with respect to the camera pose, and plan or control a trajectory for scene exploration based on this local gradient information.



\addtolength{\textheight}{-8cm}   
\pagebreak

\bibliographystyle{IEEEtran}
\bibliography{references}

\end{document}

%% file: mathstuff.tex
\newcommand{\vect}[1]{\mathbf{ #1}}
\newcommand{\vectg}[1]{{\boldsymbol{ #1}}}

\newcommand{\R}{\mathbb{R}}

\newcommand{\vI}{\vect{I}}

\newcommand{\vc}{\vect{c}}
\newcommand{\vd}{\vect{d}}

\newcommand{\vo}{\vect{o}}

\newcommand{\vr}{\vect{r}}

\newcommand{\vx}{\vect{x}}

\newcommand{\vmu}{\vectg{\mu}}

\newcommand{\vtheta}{\vectg{\theta}}

\newcommand{\vsigma}{\vectg{\sigma}}

\newcommand{\cI}{\mathcal{I}}

\newcommand{\cN}{\mathcal{N}}

%% file: uncertainty.tex
\section{Density-Aware NeRF Ensembles}

\subsection{Preliminaries}
A NeRF is a parametric function $f_\vtheta(\vx, \vd): \R^3 \times \R^2 \rightarrow \R^4$, implemented as a deep neural network with parameters $\vtheta$, that encodes the density $\rho$ and colour $\vc$ of all point-direction pairs $(\vx, \vd)$ in a scene. These density and colour predictions can be used to render a new view of the scene represented by the NeRF through the following process. Given a ray $\vr(t) = \vo + t\vd$ with origin point $\vo$ and direction vector $\vd$, the expected colour $C(\vr)$ of camera ray $\vr(t)$ with near and far bounds $t_n$ and $t_f$ is
\begin{equation}
    C(\vr) = \int_{t_n}^{t_f} T(t) \cdot \rho(\vr_{(t)}) \cdot \vc(\vr_{(t)}, \vd) \;dt
\end{equation}
In practice this integral is approximated via the quadrature rule, using discrete sums and stratified sampling from $N$ evenly spaced bins to make the rendering process tractable. Through a series of simplifications that are not relevant for the remainder of the paper (we refer the interested reader to~\cite{mildenhall2021nerf}), this becomes the convenient expression
\begin{equation}
    C(\vr) = \sum_{i=1}^N 
    \overbrace{
        \underbrace{\prod_{j=1}^{i-1} (1-o_j)}_\text{transmittance}
        \cdot 
        \underbrace{o_i}_\text{occupancy}  
    }^\text{termination probability}
    \cdot 
    \overbrace{\vc_i}^\text{colour}
    \label{eq:render}
\end{equation}

Given a neural radiance field $f_\vtheta$ and equation (\ref{eq:render}), we can render an image $\cI$ by evaluating $C(\vr)$ for all rays $\vr$ that pass through the camera center and the image plane. In the following, we use the notation $c_\vtheta(\vr)$ to indicate the predicted colour along ray $\vr$ using the rendering process of (\ref{eq:render}) and the NeRF $f_\vtheta$.

\subsection{NeRF Ensembles for Predictive RGB Uncertainty}
\label{sec:ensemble}
Following the Deep Ensembles approach~\cite{lakshminarayanan2017simple}, we propose to quantify the predictive uncertainty of a NeRF by training an \emph{ensemble} of networks $\{f_{\vtheta_k}\}_{k=1...M}$. The $M$ ensemble members are initialised with different parameters $\vtheta^{(0)}_k$, but trained on the same data. By interpreting the ensemble as a uniformly-weighted mixture model, the members' predictions are combined through averaging, and the predictive uncertainty is expressed as the variance over the individual member predictions.
With an ensemble of NeRFs, the expected colour of ray $\vr$ in a scene is
\begin{equation}
\vmu_\text{RGB}(\vr) = \frac{1}{M}\sum_{k=1}^M c_{\vtheta_k}(\vr).
\end{equation}
The predictive uncertainty can be expressed as the variance over the individual member predictions: 
\begin{equation}
    \vsigma_\text{RGB}^2(\vr) = \frac{1}{M}\sum_{k=1}^M \left(\vmu(\vr) - c_{\vtheta_k}(\vr)\right)^2.
\end{equation}
$\vmu_\text{RGB}$ and $\vsigma_\text{RGB}^2$ can be calculated very easily by rendering the $M$ individual RGB images $\cI_i$ and calculating the mean and variance directly in pixel space. Both will be 3-vectors over the RGB colour channels, i.e we do not consider the covariance \emph{between} colour channels.

We combine the variances from the colour channels into a single variance by taking the average along the three channels:
\begin{equation}
    \bar\sigma^2_\text{RGB}(\vr) = \frac{1}{3} \cdot \sum_{c \in \{RGB\}} \vsigma^2_{\text{RGB}, (c)}(\vr), 
    \label{eq:sigma_rgb}
\end{equation}
where $\vsigma^2_{\text{RGB}, (c)}(\vr)$ indicates the variance associated with colour channel $c$.

\subsection{Limitations of Simple Ensembling}
Ensembling NeRFs and using the variance in RGB space to quantify the predictive uncertainty (i.e. aleatoric and epistemic) is a simple and partially effective method. However, this approach can fail to capture the model's epistemic uncertainty arising from parts of the scene that have \emph{not} been observed during training. 

Fig.~\ref{fig:hero} illustrates an instructive example; when rendering a novel view that exposes parts of the floor, the NeRF is forced to render areas of the scene that have never been observed during training. Although one would expect the NeRF ensemble to express high uncertainty in these image regions, all ensemble members agree to render this area in black, resulting in negligible variance in colour space $\bar\sigma^2_\text{RGB}$.

\subsection{Density-Aware Ensembles to Capture Epistemic Uncertainty in Unseen Areas}
A closer inspection of the ensemble predictions along a ray $\vr$ in previously unobserved regions reveals that the individual NeRFs assign a low termination probability along all sample points on $\vr$.
The sum of the termination probabilities along the ray is close to zero (see Fig.~\ref{fig:rays} (left)), indicating the model does not assign belief to the hypothesis that the ray intersects the scene geometry within the near and far rendering bounds $t_n$ and $t_f$. In short, writing the sum of the termination probabilities along a ray $\vr$ as $q_{\vtheta_k}(\vr)$, we observe
\begin{equation}
    q_{\vtheta_k}(\vr) = \sum_{i=1}^N 
    \overbrace{
        \underbrace{\prod_{j=1}^{i-1} (1-o_j)}_\text{transmittance}
        \cdot 
        \underbrace{o_i}_\text{occupancy}  
    }^\text{termination probability at sample $i$}
    \approx 0.    
    \label{eq:sum_termination}
\end{equation}
The average summed termination probability along ray $\vr$ across the ensemble is then given by
\begin{equation}
    \bar{q}(\vr)  =  \frac{1}{M}\sum_{k=1}^M q_{\vtheta_k}(\vr),
\end{equation}
where $\bar{q}(\vr)\approx 1$ for rays that intersect with the scene structure observed during training, and $\bar{q}(\vr)\approx 0$ otherwise.
We interpret this as an expression of some of the model's \emph{epistemic} uncertainty, arising from a fundamental lack of knowledge about the scene geometry and appearance along $\vr$.
To capture this uncertainty, we introduce an additional density-aware epistemic term $\sigma^2_\text{epi}(\vr)$ that we define as:
\begin{equation}
    \sigma^2_\text{epi}(\vr) = \left(1-\bar{q}(\vr)\right)^2
\end{equation}

Finally, we combine the uncertainty measures based on the RGB-variance and above epistemic measure into the overall uncertainty $\psi^2(\vr)$:
\begin{equation}
    \psi^2(\vr) = \bar\sigma^2_\text{RGB}(\vr) + \sigma^2_\text{epi}(\vr)
\end{equation}

The predicted rendered colour along a ray according to the ensemble is then modelled as a Gaussian with diagonal covariance matrix:
\begin{equation}
    \tilde C(\vr) \sim \cN\left(\vmu_\text{RGB}(\vr), \vI_{3\times 3} \cdot \psi^2(\vr) \right)
\end{equation}

We call this method \emph{Density-aware} Ensembling, as $\sigma^2_\text{epi}(\vr)$ depends on the individual density predictions along each ray. As the experiments in the next section will show,  $\sigma^2_\text{epi}(\vr)$ and $\bar\sigma^2_\text{RGB}(\vr)$ are complementary, capturing different aspects of the model's aleatoric and epistemic uncertainty to different.
    




\section{Experiments and Results}
\begin{figure*}[t]
    \centering
    \includegraphics[width=0.24\linewidth]{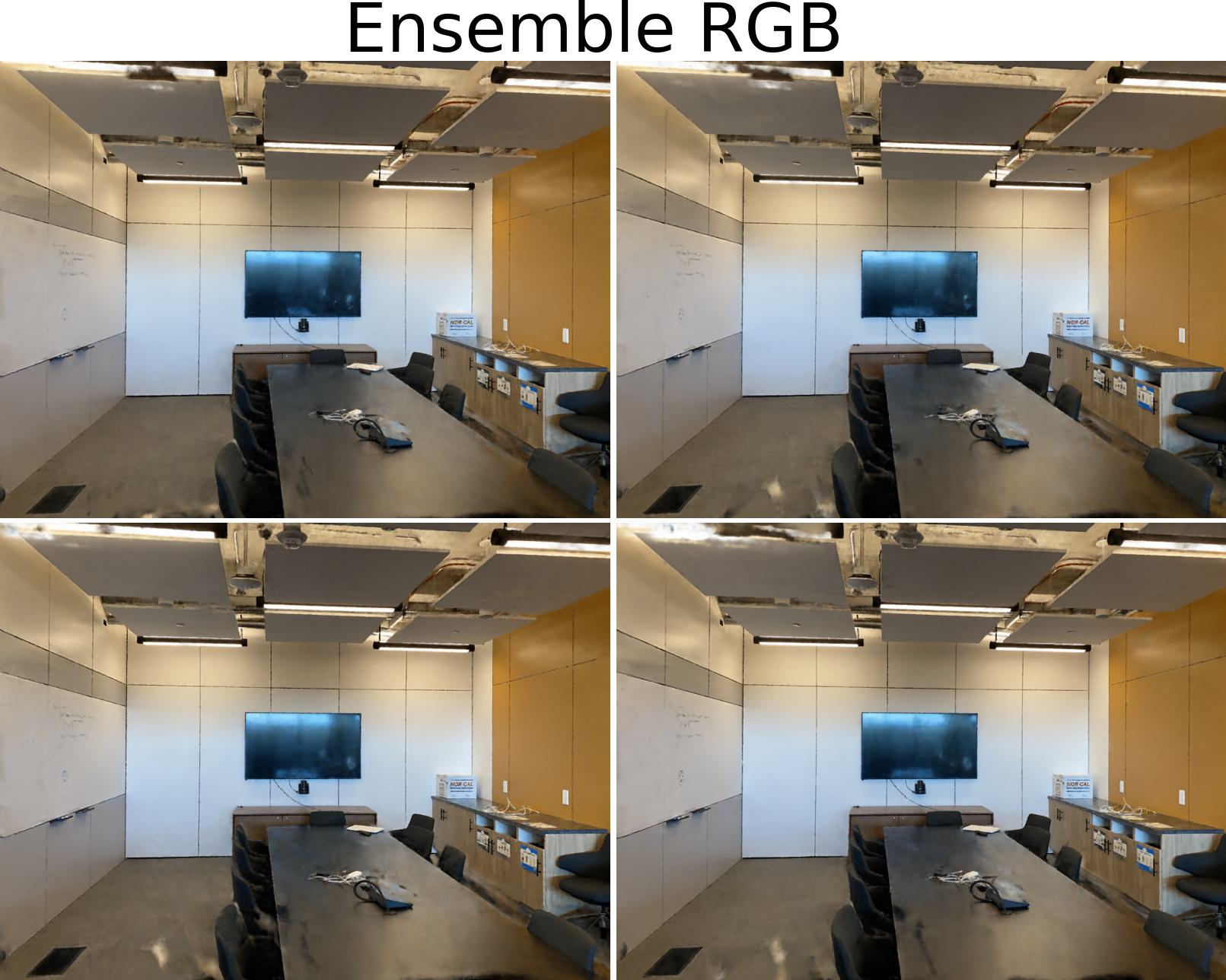} 
    \includegraphics[width=0.24\linewidth]{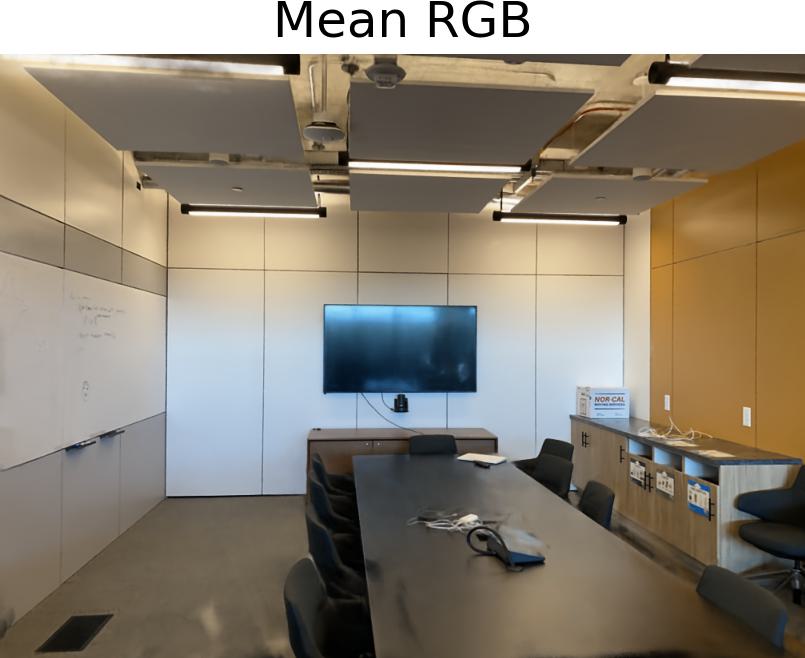} 
    \includegraphics[width=0.24\linewidth]{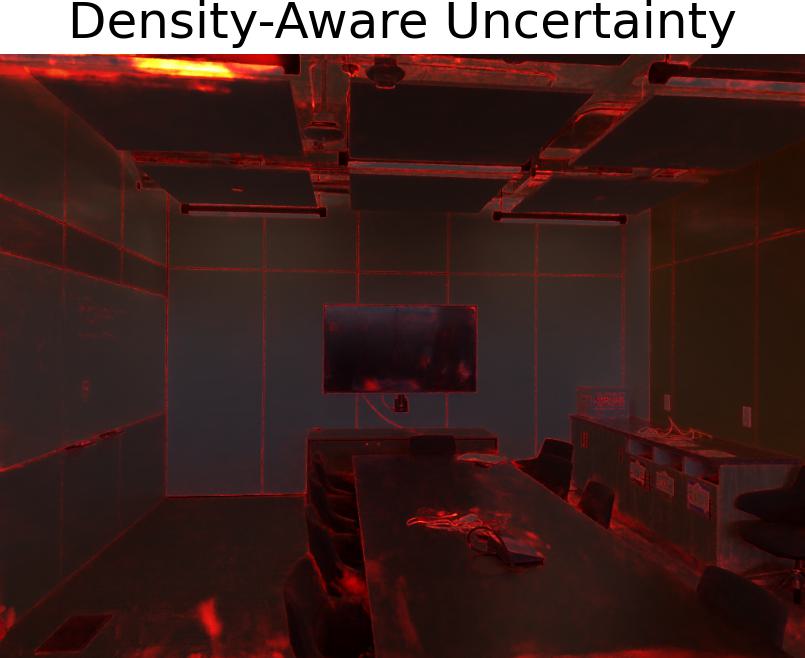} 
    \includegraphics[width=0.24\linewidth]{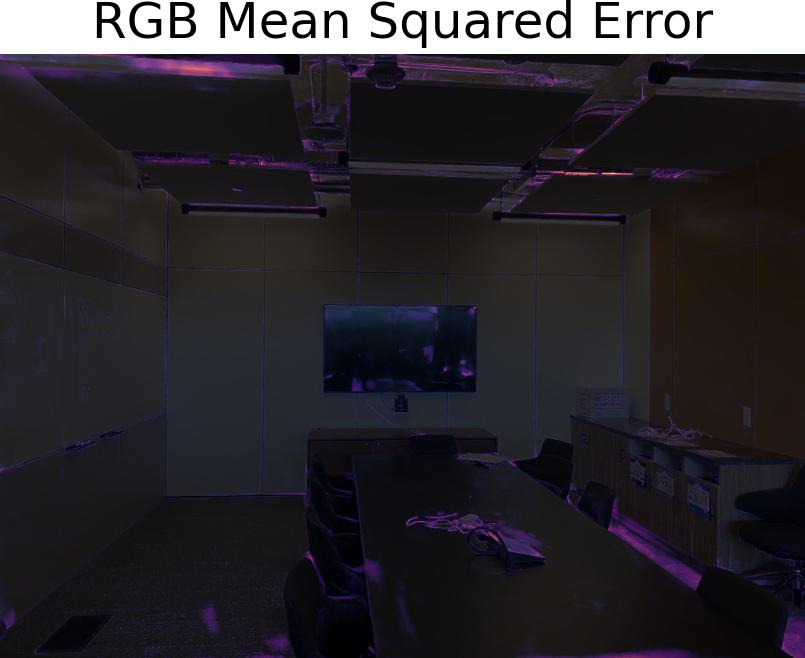}     
    \includegraphics[width=0.24\linewidth]{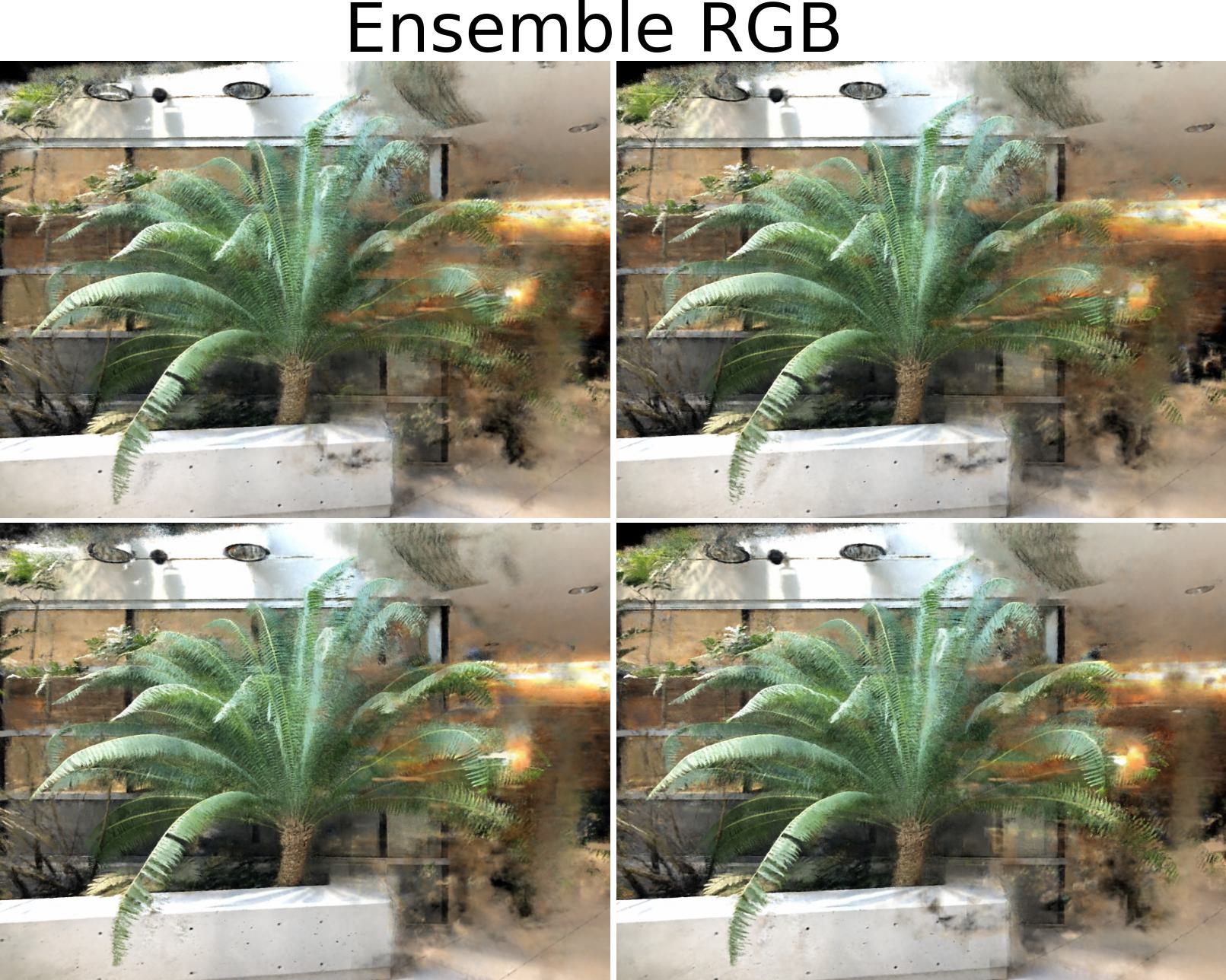} 
    \includegraphics[width=0.24\linewidth]{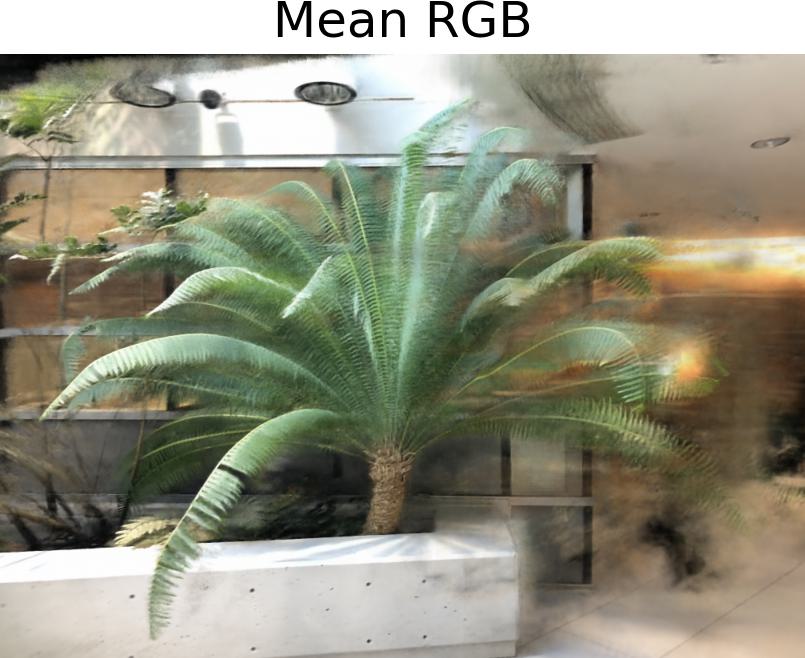} 
    \includegraphics[width=0.24\linewidth]{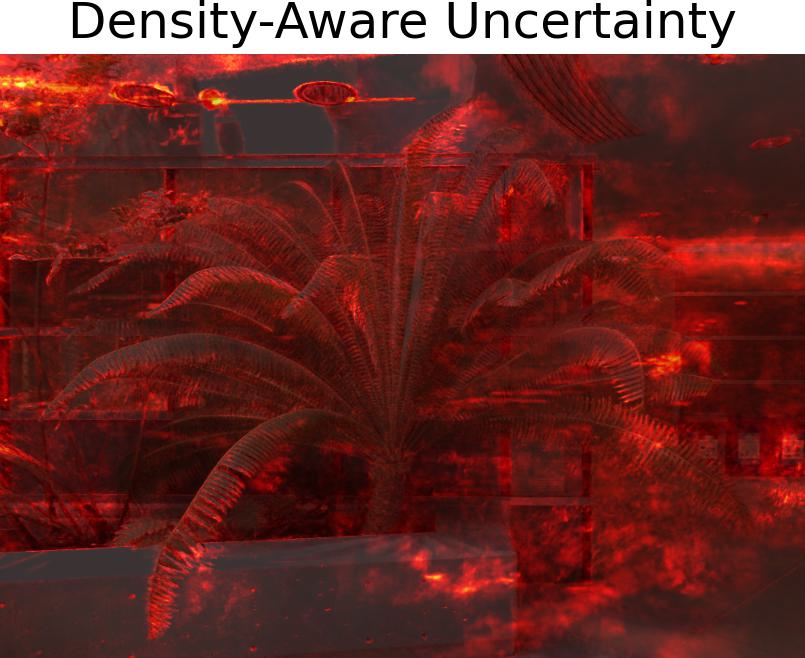} 
    \includegraphics[width=0.24\linewidth]{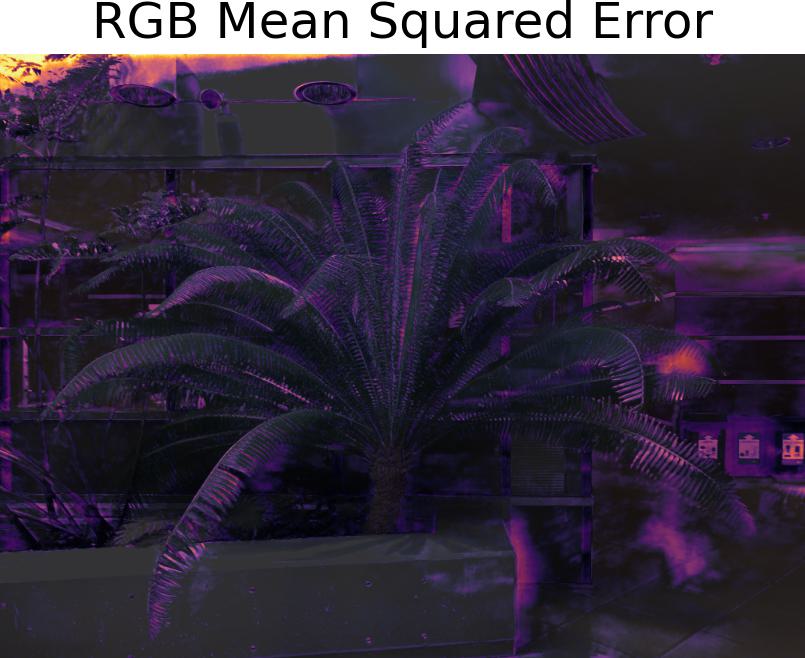}   
    \caption{Qualitative results for two views of the \emph{Room} and \emph{Fern} scene of the LLFF dataset.}
    \label{fig:qualitative}
\end{figure*}

\begin{table*}[tb]
    \centering
    \caption{Measuring uncertainty quantification with Negative Log Likelihood (NLL) for baselines and different ensemble sizes of our proposed method. 
    Considered baselines are Monte Carlo Dropout (MC-DO), a naive ensembles approach implemented by~\cite{shen2021stochastic}, NeRF in the Wild (NeRF-W)~\cite{martinbrualla2020nerfw}, S-NeRF~\cite{shen2021stochastic}, and CF-NeRF~\cite{shen2022conditional}. The latter paper only published average NLL over all scenes instead of individual results. A $\dagger$ indicates results taken from~\cite{shen2021stochastic}, $\ddagger$ from~\cite{shen2022conditional}.}
    \begin{tabular}{@{}rc|cccccc|rrrr@{}}\toprule
      & Training & \multicolumn{6}{c}{Negative Log-Likelihood $\downarrow$} 
      & \multicolumn{4}{c}{Ablation Ensemble Size (NLL $\downarrow$)} \\
     & Views & MC-DO$\dagger$ & Naive Ens$\dagger$ & NeRF-W$\dagger$ & S-NeRF$\dagger$ & CF-NeRF$\ddagger$ & Ours  &  &  & \\   
    Dataset & (Ours) & $M=5$ & $M=5$ & \cite{martinbrualla2020nerfw} & \cite{shen2021stochastic} & \cite{shen2022conditional} & $M=5$ & $M=2$ & $M=4$ & $M=8$ & $M=10$ \\
    \midrule    
    Flower & 7 & 4.63 & 1.63 & 1.71 & 1.27 & -- & \textbf{1.00} & 1.88 & 1.13 & 0.90 & 0.85\\
    Fortress & 8 & 5.19 & 2.29 & 1.04 & -0.03 & -- & \textbf{-1.30} & -1.28 & -1.29 & -1.30 & -1.30\\
    Leaves & 5 & 2.72 & 2.66 & 0.79 & \textbf{0.68} & -- & 0.97 & 2.32 & 1.10 & 0.80 & 0.73\\
    Horns & 12 & 4.18 & 2.17 & 0.78 & 0.60 & -- & \textbf{-0.55} & 0.06 & -0.50 & -0.64 & -0.66 \\
    T-Rex & 11 & 4.10 & 2.28 & 1.91 & 1.37 & -- & \textbf{-0.31} & 2.69 &  0.00 & -0.65 & -0.69\\ 
    Fern & 4 & 4.90 & 2.47 & 2.16 & 2.01 & -- & \textbf{-0.98} & -0.89 & -0.97 & -0.99 & -1.00  \\
    Orchids & 5 & 5.74 & 2.23 & 2.24 & 1.95 & -- & \textbf{-0.28} & 0.06 & -0.17 & -0.29 & -0.31\\
    Room & 8 & 5.06 & 2.13 & 4.93 & 2.35 & -- & \textbf{-1.35} &  -1.29 & -1.34 & -1.35 & -1.35 \\ \midrule
    Average & & 4.57 & 2.23 & 1.95 & 1.27 & 0.57 & \textbf{-0.35} & 0.44 & -0.26 & -0.44 & -0.47 \\    
    \bottomrule
    \end{tabular}   
    \label{tab:main_results}
\end{table*}

\begin{table*}[tb]
    \centering
    \caption{Ablation study on the influence of the individual components of the uncertainty measure $\psi^2$. We report the average-mean and average-median NLL per scene along with the standard deviations. See the text for explanation.}
     \begin{tabular}{@{}r|cc|cc|cc@{}}\toprule
        & \multicolumn{6}{c}{Negative Log-Likelihood $\downarrow$ ($M=10$)} \\ 
        & \multicolumn{2}{c}{$\psi^2 = \sigma^2_\text{RGB} + \sigma^2_\text{epi}$}
        & \multicolumn{2}{c}{$\psi^2 = \sigma^2_\text{RGB} $}
        & \multicolumn{2}{c}{$\psi^2 = \sigma^2_\text{epi}$} \\       
        Dataset & Mean & Median & Mean & Median & Mean & Median  \\         
        \midrule
        Flower  & 0.85 $\pm$ 0.26 & -0.07 $\pm$ 0.18 & 1997 $\pm$ 2764 & 0.76 $\pm$ 0.39 & 2.85 $\pm$ 0.95 & 0.46 $\pm$ 0.38 \\
        Fortress & -1.30 $\pm$ 0.07 & -1.40 $\pm$ 0.01 & -0.51 $\pm$ 1.17 & -2.26 $\pm$ 0.35 & -1.26 $\pm$ 0.10 & -1.42 $\pm$ 0.01\\
        Leaves & 0.73 $\pm$ 0.174 & -0.20 $\pm$ 0.10 & 3762 $\pm$ 3821 & -0.06 $\pm$ 0.15 & 4.76 $\pm$ 1.72 & 0.72 $\pm$ 0.32\\
        Horns & -0.66 $\pm$ 0.26 & -1.35 $\pm$ 0.10 & 3.08 $\pm$ 2.81 & -1.49 $\pm$ 0.26 & 0.35 $\pm$ 0.60 & -1.42 $\pm$ 0.08 \\
        T-Rex & -0.69 $\pm$ 0.53 & -1.50 $\pm$ 0.05 & 137 $\pm$ 480 & -1.62 $\pm$ 0.821 & 7.36 $\pm$ 2.62 & -1.49 $\pm$ 0.05\\
        Fern & -1.00 $\pm$ 0.11 & -1.30 $\pm$ 0.05 & 14.2 $\pm$ 39.7 & -1.66 $\pm$ 0.23 & -0.90 $\pm$ 0.14 & -1.39 $\pm$ 0.02\\
        Orchids & -0.31 $\pm$ 0.09 & -0.95 $\pm$ 0.06 & 1.51 $\pm$ 0.55 & -0.911 $\pm$ 0.11 & 0.51 $\pm$ 0.26 & -1.09 $\pm$ 0.09\\
        Room & -1.35 $\pm$ 0.08 & -1.43 $\pm$ 0.01 & -0.54 $\pm$ 1.51 & -2.58 $\pm$ 0.20 & -1.30 $\pm$ 0.10 & -1.44 $\pm$ 0.01   \\ \midrule
        Average & -0.47 & -1.02 &  739 & -1.23 & 1.55 & -0.88 \\            
        \bottomrule
     \end{tabular}
    \label{tab:ablation}
\end{table*}

\subsection{Experimental Setup}
\noindent\textbf{Datasets:} We follow the evaluation protocol for uncertainty quantification in NeRFs established by~\cite{shen2021stochastic}, evaluating our approach on the eight scenes of the LLFF dataset~\cite{mildenhall2021nerf} -- 3 outdoor scenes (\emph{Flower, Leaves, Orchids}) and 5 indoor scenes (\emph{Fortress, Horns, T-Rex, Room, Fern}). As motivated in~\cite{shen2021stochastic}, we randomly split the available views from the individual datasets into 20\% for training, and keep the remaining 80\% for testing. This way, we can evaluate our model in a scenario where only a few (4-12) training views of a scene are available, which is more realistic for robotics applications than the dense viewpoint coverage typically encountered in NeRF datasets.

\noindent\textbf{Baselines:} Using the established datasets for evaluation allows us to directly compare with S-NeRF~\cite{shen2021stochastic} and the baseline results published in~\cite{shen2021stochastic}. These are Monte Carlo Dropout sampling~\cite{gal2016dropout}, a naive ensembling approach based on deep ensembles~\cite{lakshminarayanan2017simple}, and NeRF in the Wild (NeRF-W)~\cite{martinbrualla2020nerfw}.

For the Monte Carlo Dropout baseline, \cite{shen2021stochastic} added a Dropout layer after every odd layer in the network, sampled 5 times, and used the variance in RGB space as the uncertainty measure. The Naive Ensembling baseline also uses the RGB variance, after training an ensemble with 5 members.
The NeRF-W experiment in \cite{shen2021stochastic} was conducted by removing the latent embedding components and keeping only the uncertainty estimation layers. We refer the reader to~\cite{martinbrualla2020nerfw} for details on the architecture of NeRF-W.

We additionally compare against CF-NeRF~\cite{shen2022conditional}, new work by the authors of~\cite{shen2021stochastic}. Since they follow the same evaluation protocol, we can directly compare against the results reported in~\cite{shen2022conditional}.

\noindent\textbf{Evaluation Metric:}
We report the Negative Log-Likelihood (NLL) as a principled and established way of assessing the quality of predictive uncertainty. NLL measures the likelihood of the \emph{true} colour at a pixel under a Gaussian model $\cN(\vmu_\text{RGB}(\vr), \psi^2(\vr))$ formed by the \emph{predicted} mean colour $\vmu_\text{RGB}$ and our squared uncertainty measure as variance.

\noindent\textbf{Implementation:} 
We implement our Density-aware Ensembles based on the publicly available Instant-NGP~\cite{mueller2022instant} implementation. A simple modification lets us access the underlying densities along each ray during rendering. This enables us to calculate $q_{\vtheta_k}(\vr)$ as per equation (\ref{eq:sum_termination}). We train 10 ensemble members for 5,000 training steps. Our machine with a Nvidia RTX 3090 trains for around 26 seconds per ensemble member, but we note that the process could be effectively parallelised. To calculate $\bar\sigma^2_\text{RGB}$, we render novel views at the full resolution of the ground truth image and calculate mean and variance in pixel space (see Section~\ref{sec:ensemble}).


\subsection{Results and Ablation Study}

\noindent \textbf{Main Results:}
We show the main results of our experiments in Table~\ref{tab:main_results}. Our Density-aware Ensemble with 5 members achieves a better NLL on average across all scenes of the LLFF dataset. While S-NeRF~\cite{shen2021stochastic} and CF-NeRF~\cite{shen2022conditional} report average performance of $1.27$ and $0.57$, our ensemble sets a new state of the art with a NLL of $-0.35$. We achieve a lower NLL on all individual scenes apart from \emph{Leaves}.

\noindent\textbf{Influence of Ensemble Size:}
\begin{figure}[t]
    \centering
    \includegraphics[width=1\linewidth]{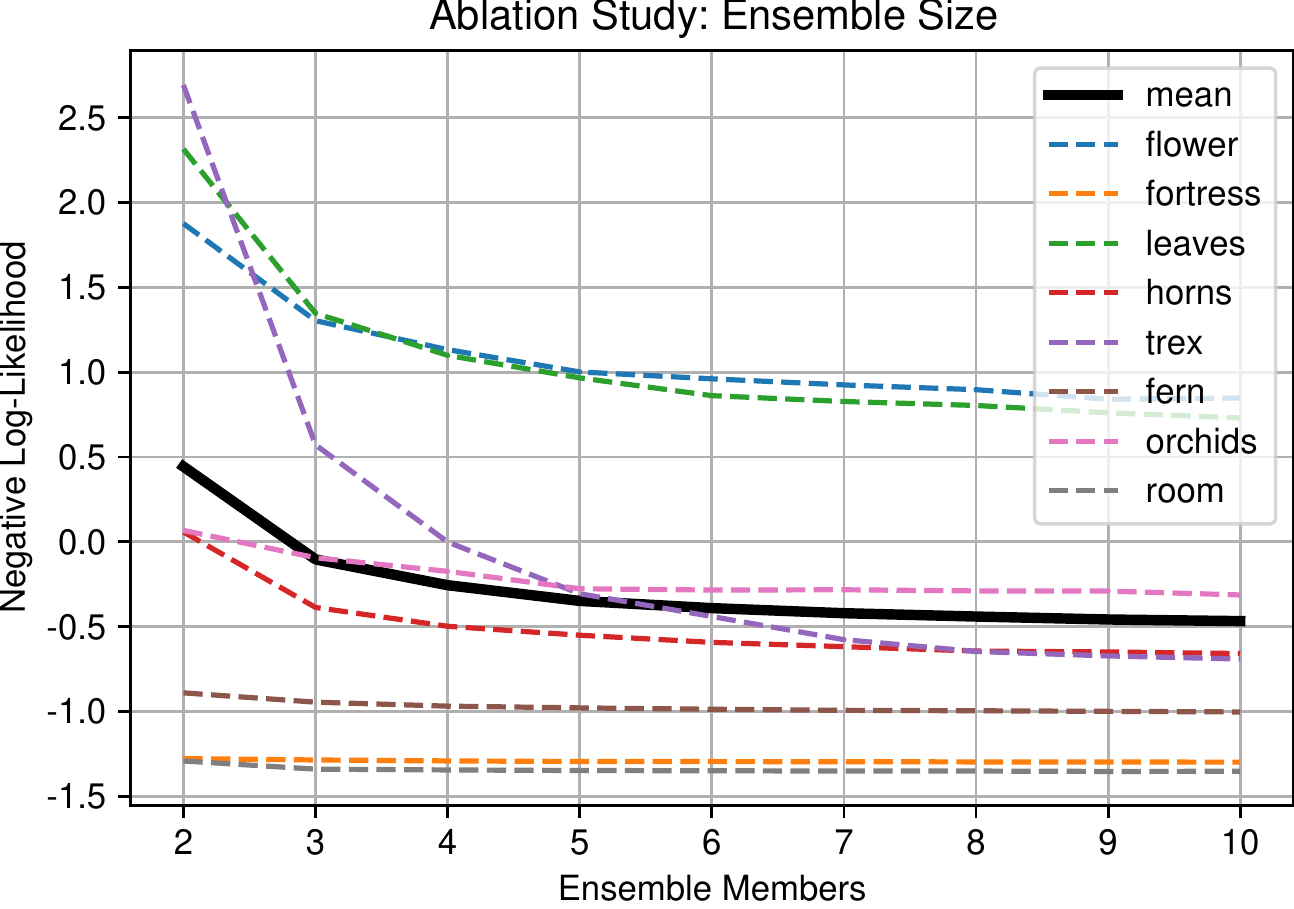}
    \caption{Ensembling is feasible and effective for moderate ensemble sizes. The Negative Log-Likelihood converges quickly for most scenes, improving negligibly beyond sizes of 5 except for \emph{T-Rex}.}
    \label{fig:ablation}
\end{figure}
We ablate the influence of $M$, the number of ensemble members. While we reported our main results for $M=5$ ensemble members to directly compare to prior work, Table~\ref{tab:main_results} also reports the results for different choices of $M$ in the four rightmost columns. As expected, a larger ensemble results in higher quality uncertainty estimates, achieving an average Negative Log-Likelihood of $-0.47$ for $M=10$ compared to $-0.35$ for the smaller ensemble with $M=5$. This is consistent with previous findings on the performance of sampling-based uncertainty quantification~\cite{lakshminarayanan2017simple, gal2016dropout}.

In Addition, Fig.~\ref{fig:ablation} plots the achieved Negative Log-Likelihood for $M=2\dots 10$. From this plot, it is apparent that although the performance increases monotonically with $M$, the gains become more and more diminishing for larger $M$. This is significant for practical use cases, as it indicates that even a relatively small NeRF ensemble can express uncertainty of high quality.

\noindent\textbf{Influence of Individual Uncertainty Measures:}
\begin{figure}[t]
    \centering
    \includegraphics[width=0.49\linewidth]{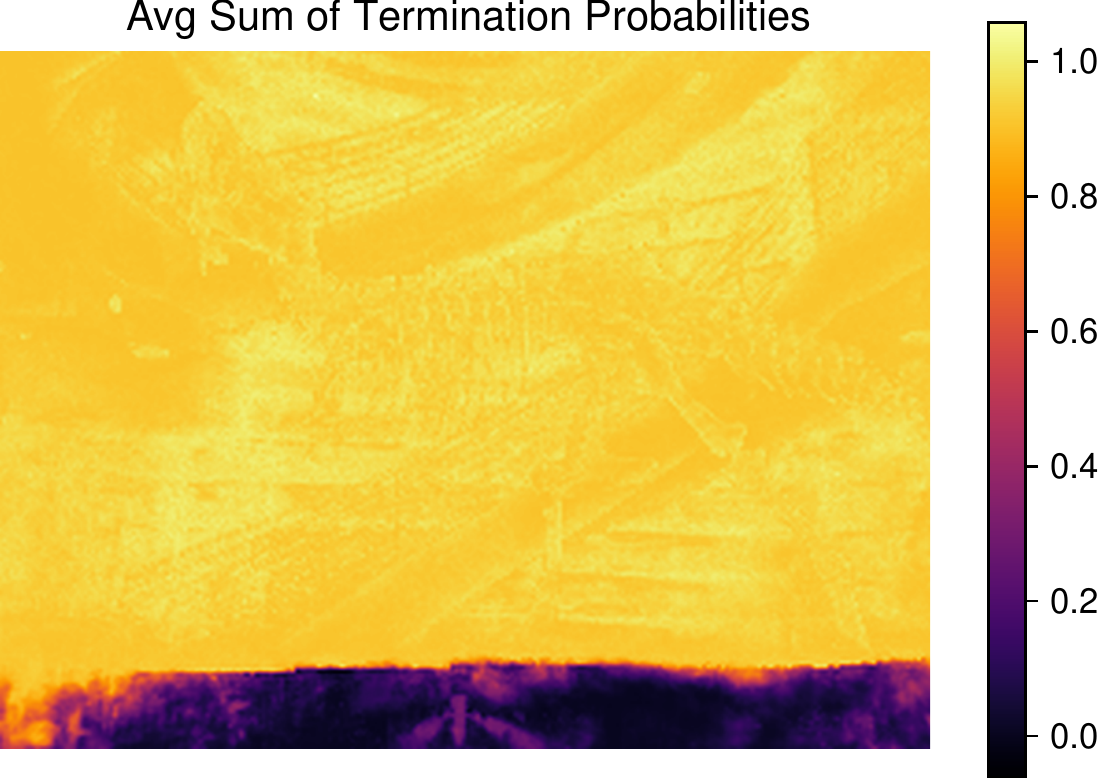} 
    \includegraphics[width=0.49\linewidth]{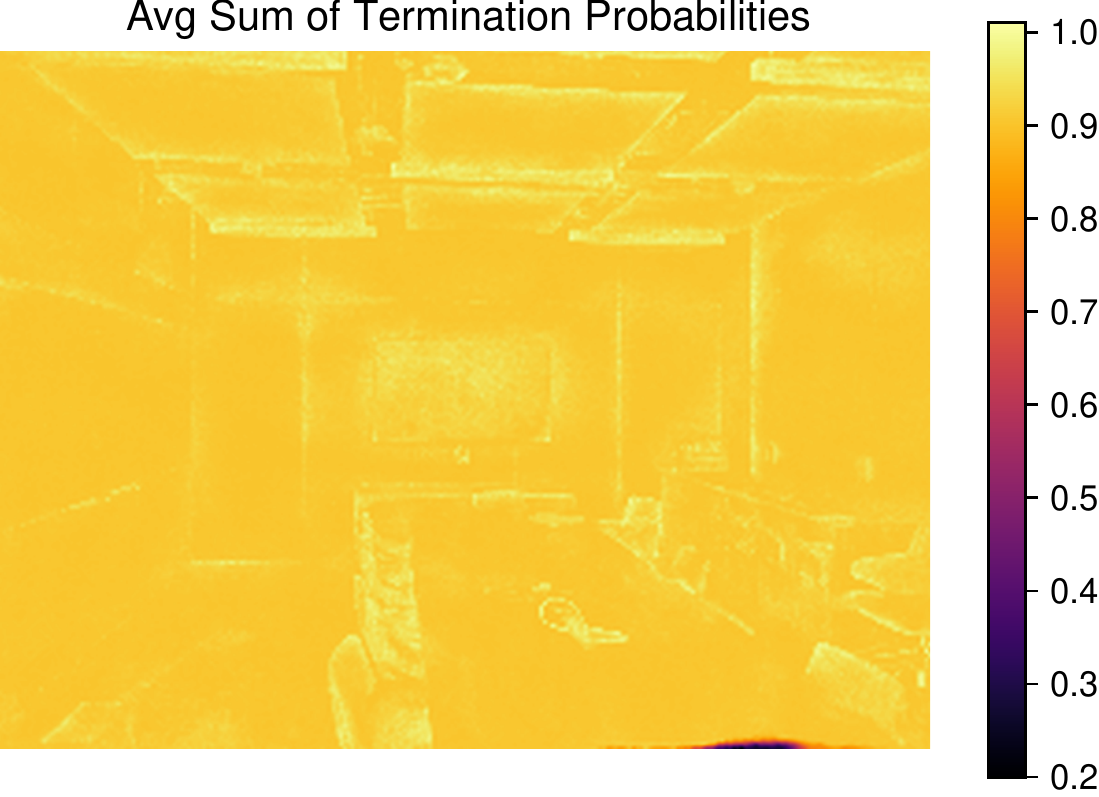} 
    \caption{The average sum of termination probabilities $\bar{q}_{\vtheta_k}(\vr)$ for a view from the \emph{T-Rex} and \emph{Room} scenes (RGB images in Fig.~\ref{fig:hero} and \ref{fig:qualitative}). The model assigns small termination probability to rays corresponding to parts of the view that were never observed in training (bottom part, left image). Interestingly, we also observe small fluctuations in termination probability per ray in other parts of the scene, where the model assigns a belief of slightly less than 1. Our ablation indicates this is correlated with the prediction error.}
    \label{fig:rays}
\end{figure}
Our proposed uncertainty measure $\psi^2(\vr)$ consists of two terms, the mean RGB variance $\bar\sigma^2_\text{RGB}$ and the epistemic variance $\sigma^2_\text{epi}$. To better understand their individual influence, Table~\ref{tab:ablation} reports results when using only either term, or both terms combined. 

For this ablation study, we report both the mean-average and the mean-median Negative Log-Likelihood, along with their standard deviations, for every scene in the LLFF datset. The mean-average NLL is calculated by averaging the NLL over all pixels in a rendered test image, and then calculating the mean over all test images in a scene. In contrast, the mean-median NLL calculates the \emph{median} NLL over all pixels, before averaging over all test views. Reporting both metrics reveals that using only the RGB-based variance $\sigma^2_\text{RGB}$ as the uncertainty measure is affected by severe outliers, i.e. pixels with very high (bad) NLL. This becomes clear when comparing the mean-average and the mean-median performances: while the former is subject to outliers, the latter is relatively robust against outliers. As explained in the motivation for our method in Section~\ref{sec:ensemble}, the outlier pixels with very high NLL are caused by parts of the scene that were not observed during training. 

As a somewhat surprising result, we observe that the epistemic uncertainty term $\sigma^2_\text{epi}$ by itself achieves reasonable performance on most scenes. Upon closer inspection, this is caused by small fluctuations in the individual $q_{\vtheta_k}(\vr)$, the sum over the termination probabilities per ray. We observe that the model assigns slightly lower $q_{\vtheta_k}(\vr)$ for rays with higher uncertainty, i.e. close to but not exactly 1. We illustrate this in Fig.~\ref{fig:rays} for views from two scenes.

However, using the sum of $\sigma^2_\text{RGB}$ and $\sigma^2_\text{epi}$ as the proposed uncertainty measure $\psi^2(\vr)$ consistently yields the best results, indicating the complementarity of both components.

%% file: exploration.tex
\section{Using Uncertainty for Next-Best View Selection and Model Refinement}
\begin{figure}[t]
    \centering
    \includegraphics[width=0.95\linewidth]{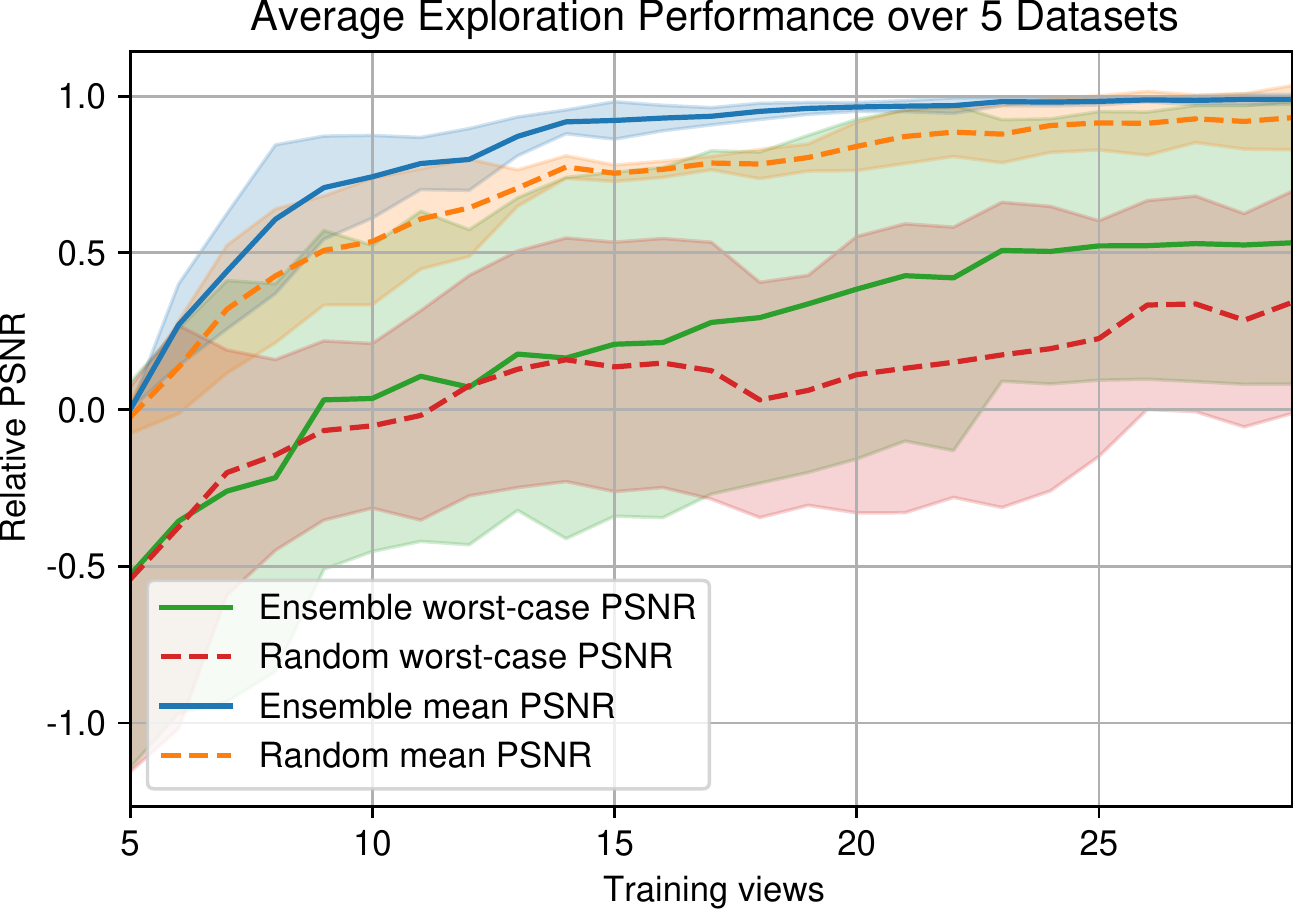}
    \caption{A NeRF ensemble can select informative next-best training views. Starting from 5 highly similar views, we incrementally add new views, chosen at random or based on the ensemble uncertainty.}
    \label{fig:exploration}
\end{figure}

\begin{figure*}[tb]
    \centering
    \includegraphics[width=0.32\linewidth]{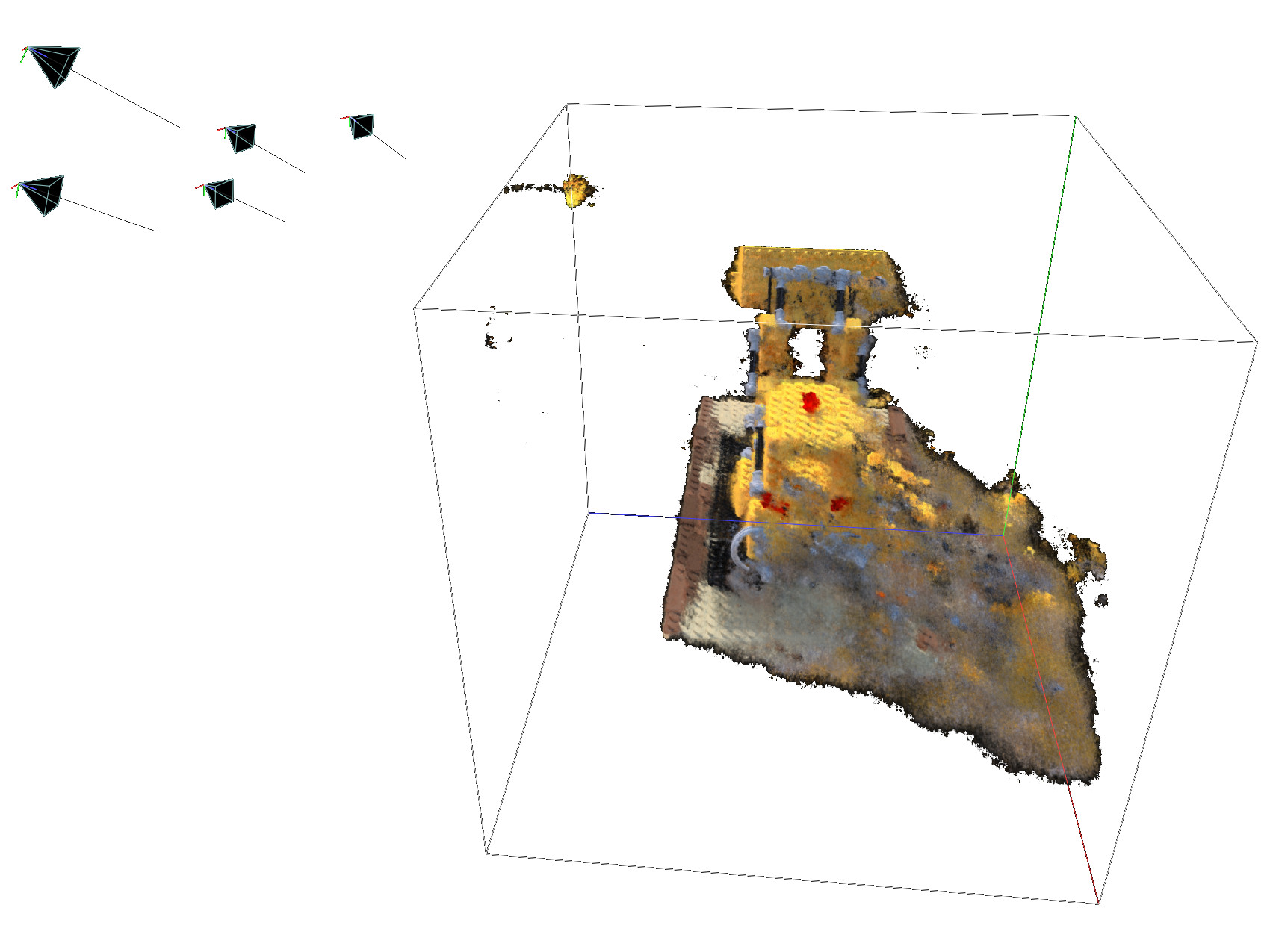} 
    \includegraphics[width=0.32\linewidth]{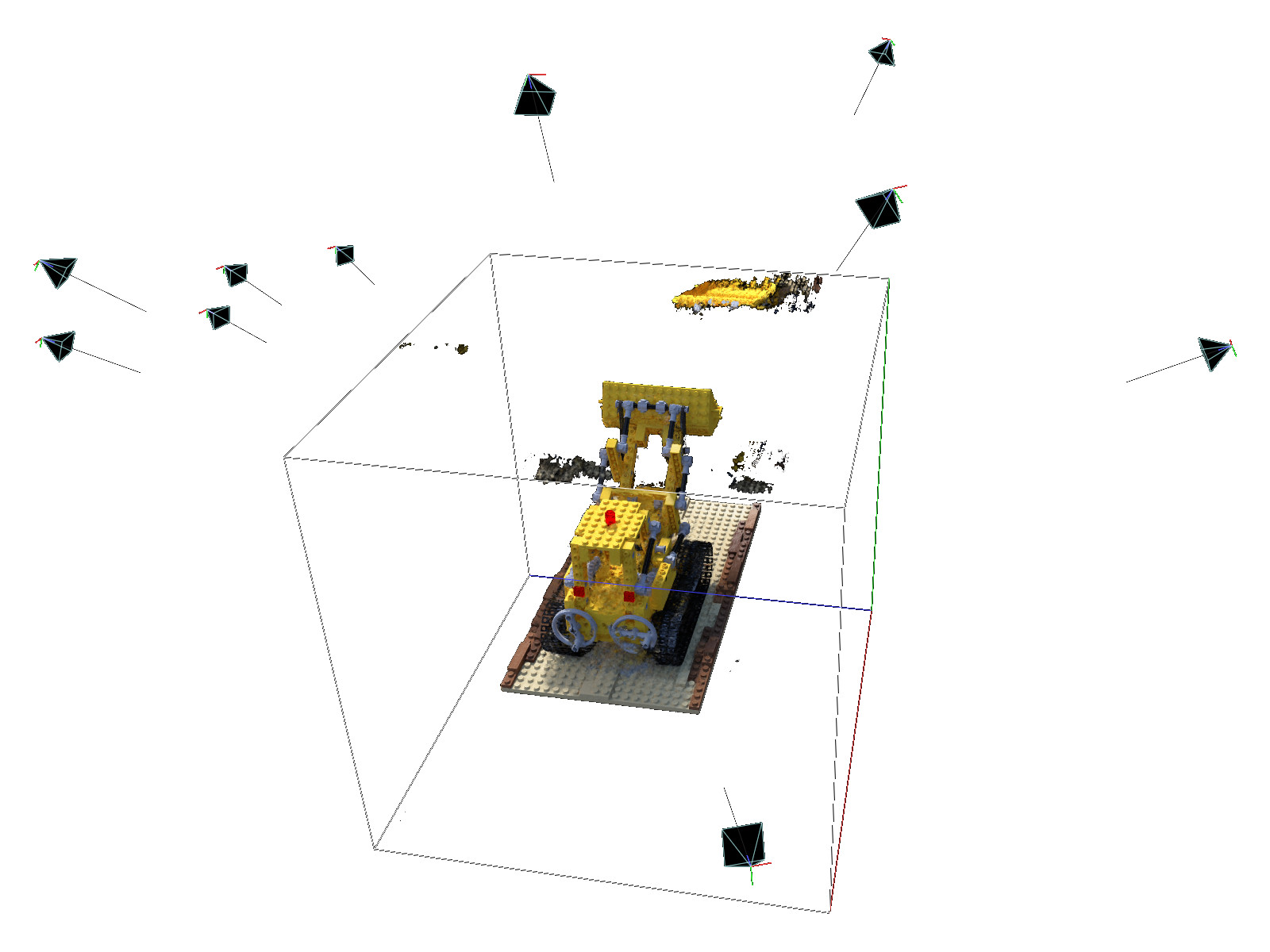}    
    \includegraphics[width=0.32\linewidth]{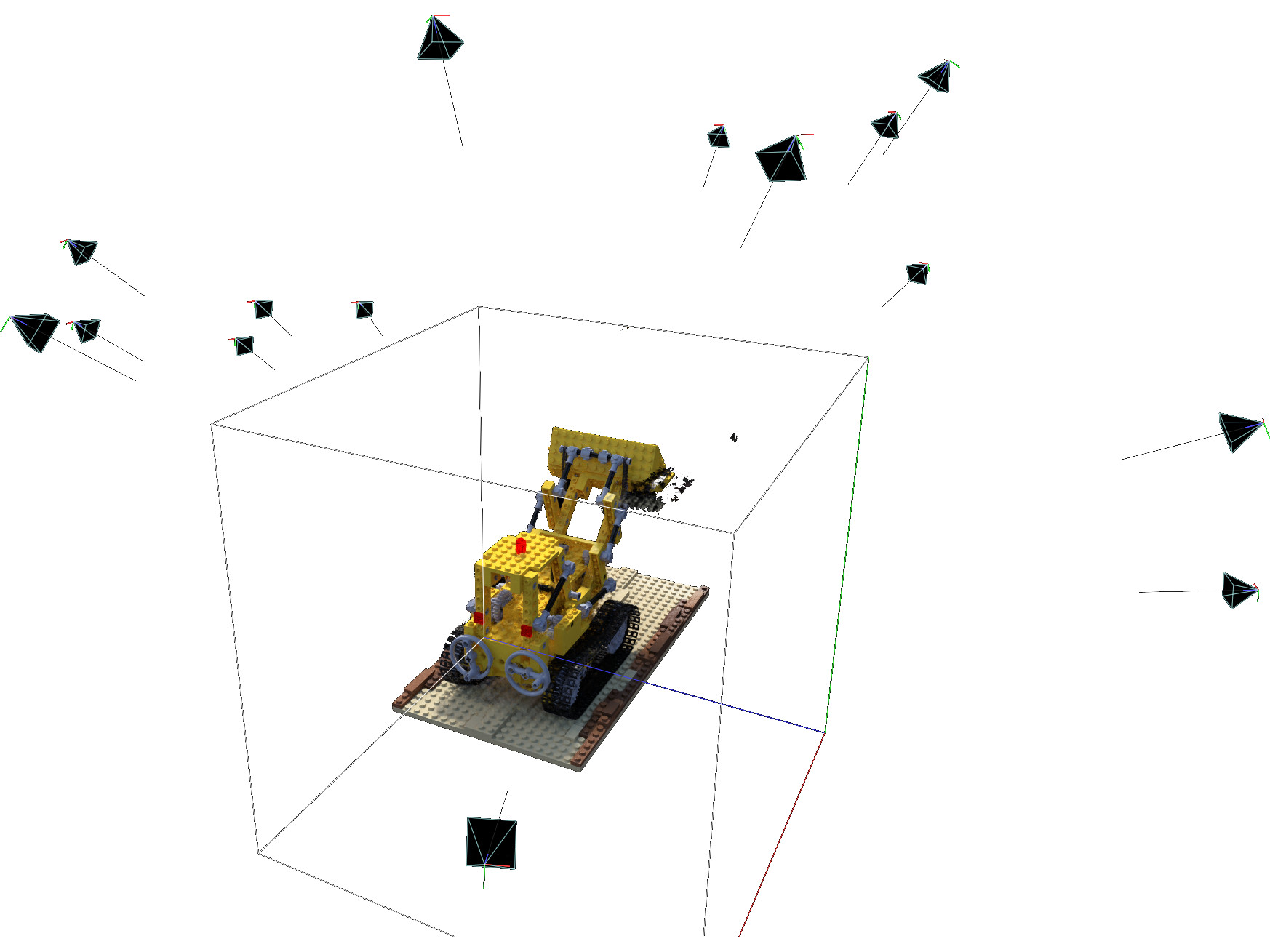}        
    \caption{Next-best view selection using ensemble uncertainty. Left: 5 initial and very similar training views lead to high model error. Centre: After selecting 5 views, the model error is significantly reduced. Right: After adding another 5 views to the training set.}
    \label{fig:exploration-qualitative}
\end{figure*}

We now describe an experiment that utilises the ensemble uncertainty for next-best view selection and model refinement.

\noindent\textbf{Datasets:} We use five synthetic scenes (\emph{Lego, Hotdog, Ficus, Drums, Microphone}) from the dataset published alongside the original NeRF paper~\cite{mildenhall2021nerf}. In these scenes, the cameras of the training set are positioned in a semi-sphere around the object, facing the object centre. From the training dataset of every scene, we randomly select 5 images from very similar viewpoints for the initial training, and keep the remaining images as candidate views to select the next-best (most informative) training image.

The test views are independent from the training data and used to evaluate the quality of the scene reconstruction, using the PSNR (Peak Signal-to-Noise Ratio) metric, as is standard in the NeRF literature~\cite{mildenhall2021nerf, shen2021stochastic}. 

\noindent\textbf{Method, Metric and Baseline:}
Starting with the initial training set of 5 views, we train an ensemble of $M=5$ NeRFs for 2,000 steps each. We then calculate the predictive uncertainty for each of the \emph{remaining} unused candidate views, and select the view with the maximum uncertainty as the next view to be added to the training data. We then retrain the NeRF and iterate this process (see Fig.~\ref{fig:exploration-qualitative}). As a simple baseline method, we select the next-best view at random.

After every training iteration, we evaluate the model quality with the independent test dataset. Comparing the rendered test views with the ground truth images, we calculate the average PSNR over all test images. We identify the lowest PSNR to measure the worst-case performance, i.e. the test view with the highest error compared to the ground truth. 

To make the PSNR comparable across all scenes, we re-scale PSNR so that the average PSNR at the beginning of our evaluation loop (i.e. only using the initial 5 views) is 0, and that the best average PSNR across all iterations (usually the PSNR of the final iteration which has access to most training images) is equal to 1. We can then plot the average and worst-case performance in one plot for both strategies of selecting the next best view (uncertainty-based selection and random selection) in Fig.~\ref{fig:exploration}. 

\noindent\textbf{Results:}
The uncertainty measured by our ensembling approach is an effective way of selecting the next-best view to add to the training dataset. As we can see in Fig.~\ref{fig:exploration}, the average-case performance is better compared to the random baseline. Although both methods eventually converge to the same performance as more images are used for training, using uncertainty for view selection achieves a higher gain in quality per training image by selecting more informative views. 

Similarly, the worst-case performance is significantly improved when selecting views using our quantified uncertainty. This indicates that our density-aware ensemble uncertainty is an effective proxy for ground-truth error -- by selecting the view with the highest uncertainty, we effectively choose to add the view that causes a high error to the training set for the next iteration.

These results are significant for robotics and active vision, where gathering exhaustive training data is too expensive or time consuming, and a trade-off between representation quality and the number of training views (or the time required to gather those views) has to be considered.

%% file: main.bbl
\begin{thebibliography}{10}
\providecommand{\url}[1]{#1}
\csname url@samestyle\endcsname
\providecommand{\newblock}{\relax}
\providecommand{\bibinfo}[2]{#2}
\providecommand{\BIBentrySTDinterwordspacing}{\spaceskip=0pt\relax}
\providecommand{\BIBentryALTinterwordstretchfactor}{4}
\providecommand{\BIBentryALTinterwordspacing}{\spaceskip=\fontdimen2\font plus
\BIBentryALTinterwordstretchfactor\fontdimen3\font minus
  \fontdimen4\font\relax}
\providecommand{\BIBforeignlanguage}[2]{{%
\expandafter\ifx\csname l@#1\endcsname\relax
\typeout{** WARNING: IEEEtran.bst: No hyphenation pattern has been}%
\typeout{** loaded for the language `#1'. Using the pattern for}%
\typeout{** the default language instead.}%
\else
\language=\csname l@#1\endcsname
\fi
#2}}
\providecommand{\BIBdecl}{\relax}
\BIBdecl

\bibitem{mildenhall2021nerf}
B.~Mildenhall, P.~P. Srinivasan, M.~Tancik, J.~T. Barron, R.~Ramamoorthi, and
  R.~Ng, ``Nerf: Representing scenes as neural radiance fields for view
  synthesis,'' \emph{Communications of the ACM}, vol.~65, no.~1, pp. 99--106,
  2021.

\bibitem{dellaert2020neural}
F.~Dellaert and L.~Yen-Chen, ``Neural volume rendering: Nerf and beyond,''
  \emph{arXiv preprint arXiv:2101.05204}, 2020.

\bibitem{adamkiewicz2022navigation}
M.~Adamkiewicz, T.~Chen, A.~Caccavale, R.~Gardner, P.~Culbertson, J.~Bohg, and
  M.~Schwager, ``Vision-only robot navigation in a neural radiance world,''
  \emph{IEEE Robotics and Automation Letters}, vol.~7, no.~2, pp. 4606--4613,
  2022.

\bibitem{sucar2021imap}
E.~Sucar, S.~Liu, J.~Ortiz, and A.~J. Davison, ``imap: Implicit mapping and
  positioning in real-time,'' in \emph{Proceedings of the IEEE/CVF
  International Conference on Computer Vision}, 2021, pp. 6229--6238.

\bibitem{zhu2022nice}
Z.~Zhu, S.~Peng, V.~Larsson, W.~Xu, H.~Bao, Z.~Cui, M.~R. Oswald, and
  M.~Pollefeys, ``Nice-slam: Neural implicit scalable encoding for slam,'' in
  \emph{Proceedings of the IEEE/CVF Conference on Computer Vision and Pattern
  Recognition}, 2022, pp. 12\,786--12\,796.

\bibitem{li2022visumotor}
Y.~Li, S.~Li, V.~Sitzmann, P.~Agrawal, and A.~Torralba, ``3d neural scene
  representations for visuomotor control,'' in \emph{Conference on Robot
  Learning}.\hskip 1em plus 0.5em minus 0.4em\relax PMLR, 2022, pp. 112--123.

\bibitem{yen2022nerf}
L.~Yen-Chen, P.~Florence, J.~T. Barron, T.-Y. Lin, A.~Rodriguez, and P.~Isola,
  ``Nerf-supervision: Learning dense object descriptors from neural radiance
  fields,'' \emph{arXiv preprint arXiv:2203.01913}, 2022.

\bibitem{ichnowski2021dex}
J.~Ichnowski, Y.~Avigal, J.~Kerr, and K.~Goldberg, ``Dex-nerf: Using a neural
  radiance field to grasp transparent objects,'' in \emph{Conference on Robot
  Learning}.\hskip 1em plus 0.5em minus 0.4em\relax PMLR, 2022, pp. 526--536.

\bibitem{mueller2022instant}
\BIBentryALTinterwordspacing
T.~M\"uller, A.~Evans, C.~Schied, and A.~Keller, ``Instant neural graphics
  primitives with a multiresolution hash encoding,'' \emph{ACM Trans. Graph.},
  vol.~41, no.~4, pp. 102:1--102:15, Jul. 2022. [Online]. Available:
  \url{https://doi.org/10.1145/3528223.3530127}
\BIBentrySTDinterwordspacing

\bibitem{cadena2016past}
C.~Cadena, L.~Carlone, H.~Carrillo, Y.~Latif, D.~Scaramuzza, J.~Neira, I.~Reid,
  and J.~J. Leonard, ``Past, present, and future of simultaneous localization
  and mapping: Toward the robust-perception age,'' \emph{IEEE Transactions on
  robotics}, vol.~32, no.~6, pp. 1309--1332, 2016.

\bibitem{rosen2021advances}
D.~M. Rosen, K.~J. Doherty, A.~Ter{\'a}n~Espinoza, and J.~J. Leonard,
  ``Advances in inference and representation for simultaneous localization and
  mapping,'' \emph{Annual Review of Control, Robotics, and Autonomous Systems},
  vol.~4, pp. 215--242, 2021.

\bibitem{garg2020semantics}
S.~Garg, N.~S{\"u}nderhauf, F.~Dayoub, D.~Morrison, A.~Cosgun, G.~Carneiro,
  Q.~Wu, T.-J. Chin, I.~Reid, S.~Gould \emph{et~al.}, ``Semantics for robotic
  mapping, perception and interaction: A survey,'' \emph{Foundations and
  Trends{\textregistered} in Robotics}, vol.~8, no. 1--2, pp. 1--224, 2020.

\bibitem{sunderhauf2018limits}
N.~S{\"u}nderhauf, O.~Brock, W.~Scheirer, R.~Hadsell, D.~Fox, J.~Leitner,
  B.~Upcroft, P.~Abbeel, W.~Burgard, M.~Milford \emph{et~al.}, ``The limits and
  potentials of deep learning for robotics,'' \emph{The International journal
  of robotics research}, vol.~37, no. 4-5, pp. 405--420, 2018.

\bibitem{thrun2002probabilistic}
S.~Thrun, W.~Burgard, and D.~Fox, \emph{Probabilistic Robotics}.\hskip 1em plus
  0.5em minus 0.4em\relax MIT Press, 2002.

\bibitem{abdar2021review}
M.~Abdar, F.~Pourpanah, S.~Hussain, D.~Rezazadegan, L.~Liu, M.~Ghavamzadeh,
  P.~Fieguth, X.~Cao, A.~Khosravi, U.~R. Acharya \emph{et~al.}, ``A review of
  uncertainty quantification in deep learning: Techniques, applications and
  challenges,'' \emph{Information Fusion}, vol.~76, pp. 243--297, 2021.

\bibitem{mackay1992practical}
D.~J. MacKay, ``A practical bayesian framework for backpropagation networks,''
  \emph{Neural computation}, vol.~4, no.~3, pp. 448--472, 1992.

\bibitem{neal2012bayesian}
R.~M. Neal, \emph{Bayesian learning for neural networks}.\hskip 1em plus 0.5em
  minus 0.4em\relax Springer Science \& Business Media, 2012, vol. 118.

\bibitem{gal2016dropout}
Y.~Gal and Z.~Ghahramani, ``Dropout as a bayesian approximation: Representing
  model uncertainty in deep learning,'' in \emph{International Conference on
  Machine Learning (ICML)}.\hskip 1em plus 0.5em minus 0.4em\relax PMLR, 2016,
  pp. 1050--1059.

\bibitem{lakshminarayanan2017simple}
B.~Lakshminarayanan, A.~Pritzel, and C.~Blundell, ``Simple and scalable
  predictive uncertainty estimation using deep ensembles,'' \emph{Advances in
  Neural Information Processing Systems (NeurIPS)}, vol.~30, 2017.

\bibitem{shen2021stochastic}
J.~Shen, A.~Ruiz, A.~Agudo, and F.~Moreno-Noguer, ``Stochastic neural radiance
  fields: Quantifying uncertainty in implicit 3d representations,'' in
  \emph{International Conference on 3D Vision (3DV)}, 2021.

\bibitem{shen2022conditional}
J.~Shen, A.~Agudo, F.~Moreno-Noguer, and A.~Ruiz, ``Conditional-flow nerf:
  Accurate 3d modelling with reliable uncertainty quantification,'' \emph{arXiv
  preprint arXiv:2203.10192}, 2022.

\bibitem{kendall2017uncertainties}
A.~Kendall and Y.~Gal, ``What uncertainties do we need in bayesian deep
  learning for computer vision?'' \emph{Advances in neural information
  processing systems}, vol.~30, 2017.

\bibitem{xie2022neural}
Y.~Xie, T.~Takikawa, S.~Saito, O.~Litany, S.~Yan, N.~Khan, F.~Tombari,
  J.~Tompkin, V.~Sitzmann, and S.~Sridhar, ``Neural fields in visual computing
  and beyond,'' in \emph{Computer Graphics Forum}, vol.~41, no.~2.\hskip 1em
  plus 0.5em minus 0.4em\relax Wiley Online Library, 2022, pp. 641--676.

\bibitem{tancik2020fourfeat}
M.~Tancik, P.~P. Srinivasan, B.~Mildenhall, S.~Fridovich-Keil, N.~Raghavan,
  U.~Singhal, R.~Ramamoorthi, J.~T. Barron, and R.~Ng, ``Fourier features let
  networks learn high frequency functions in low dimensional domains,''
  \emph{NeurIPS}, 2020.

\bibitem{Chen2022ECCV}
A.~Chen, Z.~Xu, A.~Geiger, J.~Yu, and H.~Su, ``Tensorf: Tensorial radiance
  fields,'' in \emph{European Conference on Computer Vision (ECCV)}, 2022.

\bibitem{yu_and_fridovichkeil2021plenoxels}
{Sara Fridovich-Keil and Alex Yu}, M.~Tancik, Q.~Chen, B.~Recht, and
  A.~Kanazawa, ``Plenoxels: Radiance fields without neural networks,'' in
  \emph{CVPR}, 2022.

\bibitem{barron2021mipnerf}
J.~T. Barron, B.~Mildenhall, M.~Tancik, P.~Hedman, R.~Martin-Brualla, and P.~P.
  Srinivasan, ``Mip-nerf: A multiscale representation for anti-aliasing neural
  radiance fields,'' 2021.

\bibitem{22-driess-NeRF-RL-preprint}
D.~Driess, I.~Schubert, P.~Florence, Y.~Li, and M.~Toussaint, ``Reinforcement
  learning with neural radiance fields,'' \emph{arXiv preprint
  arXiv:2206.01634}, 2022.

\bibitem{martinbrualla2020nerfw}
R.~Martin-Brualla, N.~Radwan, M.~S.~M. Sajjadi, J.~T. Barron, A.~Dosovitskiy,
  and D.~Duckworth, ``{NeRF in the Wild: Neural Radiance Fields for
  Unconstrained Photo Collections},'' in \emph{CVPR}, 2021.

\end{thebibliography}
